%% file: root.tex
\documentclass[10pt, conference, compsocconf]{IEEEtran}

\IEEEoverridecommandlockouts                              %

\input{preamble.tex}

\makeatletter
\usepackage[utf8]{inputenc}
\usepackage{commath}
\usepackage{mathtools}
\usepackage{mathabx}
\usepackage{siunitx}
\usepackage[ruled,vlined]{algorithm2e}
\usepackage[lofdepth,lotdepth]{subfig}

\let\oldtheequation\theequation
\renewcommand\tagform@[1]{\maketag@@@{\ignorespaces#1\unskip\@@italiccorr}}
\renewcommand\theequation{(\oldtheequation)}
\makeatother

\usepackage[belowskip=0pt,aboveskip=1pt]{caption}  %
\graphicspath{{media/}}

\newcommand{\C}{\bm{C}}

\newcommand{\J}{\bm{J}}
\newcommand{\E}{\mathbb{E}}
\newcommand{\I}{\bm{1}}

\acrodef{lidar}{Light Detection And Ranging}
\acrodefplural{lidar}{Light Detection And Ranging}
\acrodef{ICP}{Iterative Closest Point}
\acrodef{GNSS}{Global Navigation Satellite System}
\acrodef{IMU}{Inertial Measurement Unit}
\acrodef{DOF}{Degrees Of Freedom}
\acrodef{RTK}{Real Time Kinematics}
\acrodef{GPS}{Global Positioning System}
\acrodef{SVD}{Singular Value Decomposition}
\acrodef{UAV}{Unmanned Aerial Vehicle}
\acrodef{IIR}{Infinite Impulse Response}
\acrodef{CAD}{Computer-aided design}

\begin{document}
\title{Accurate outdoor ground truth based on total stations}
\author{\IEEEauthorblockN{Maxime Vaidis, Philippe Gigu\`{e}re, François Pomerleau, Vladim\'ir Kubelka}
\IEEEauthorblockA{Northern Robotics Laboratory, Université Laval, Québec City, Québec, Canada\\
$\{$maxime.vaidis, philippe.giguere, francois.pomerleau, vladimir.kubelka$\}$@norlab.ulaval.ca}
}

\linepenalty=3000
\addtolength{\textfloatsep}{-0.1in}

\maketitle
\thispagestyle{plain}
\pagestyle{plain}

\begin{abstract}

In robotics, accurate ground-truth position fostered the development of mapping and localization algorithms through the creation of cornerstone datasets.
In outdoor environments and over long distances, total stations are the most accurate and precise measurement instruments for this purpose.
Most total station-based systems in the literature are limited to three \acp{DOF}, due to the use of a single-prism tracking approach.
In this paper, we present preliminary work on measuring a full pose of a vehicle, bringing the referencing system to six \acp{DOF}.
Three total stations are used to track in real time three prisms attached to a target platform.
We describe the structure of the referencing system and the protocol for acquiring the ground truth with this system.
We evaluated its precision in a variety of different outdoor environments, ranging from open-sky to forest trails, and compare this system with another popular source of reference position, the \ac{RTK} positioning solution.
Results show that our approach is the most precise, reaching an average positional error of 10$\,$mm and 0.6$\,$deg.
This difference in performance was particularly stark in environments where \ac{GNSS} signals can be weaker due to overreaching vegetation.
\end{abstract}


\IEEEpeerreviewmaketitle

\acresetall
\section{Introduction}
\input{tex/intro}

\section{Related work}
\label{sec:related_work}
\input{tex/RW}

\section{Referencing system design}
\label{sec:theory}
\input{tex/theory}

\section{Experiments}
\label{sec:experiments}

\input{tex/experimentations}

\section{Conclusion}
\input{tex/conclusion}

\section*{Acknowledgment}

This  research  was  supported  by  the  Natural  Sciences  and Engineering  Research  Council  of  Canada  (NSERC)  through the grant CRDPJ 527642-18 SNOW (Self-driving Navigation Optimized for Winter).

\IEEEtriggeratref{6}
\IEEEtriggercmd{\enlargethispage{-0.1in}}

\printbibliography

\end{document}

%% file: preamble.tex
\usepackage[l2tabu,orthodox]{nag}

\usepackage[
backend=biber, 
bibencoding=utf8,
style=ieee, 
sorting=none, 
natbib=true, 
doi=false, 
isbn=false, 
url=false, 
eprint=false, 
maxcitenames=1, 
mincitenames=1,
]{biblatex}

\usepackage[draft,pdftex,colorlinks]{hyperref}
\usepackage[printonlyused]{acronym}

\usepackage{siunitx}
\usepackage[all]{nowidow}

\usepackage[dvipsnames]{xcolor}

\usepackage{lipsum}

\usepackage[pdftex]{graphicx}

\usepackage{epstopdf}

\usepackage{import}

\graphicspath{{./latexGoodPractices/}}

\usepackage{booktabs}

\usepackage{tabu}

\usepackage{amssymb,amsfonts,amsmath,amscd}

\usepackage{bm}

\newcommand{\bbm}{\begin{bmatrix}}
\newcommand{\ebm}{\end{bmatrix}}

\usepackage[normalem]{ulem} %

%% file: tex/intro.tex
Access to precise and accurate ground-truth pose reference during experiments is vital in mobile robotics.
For instance, knowing the pose of a robot, that is its position and orientation, facilitates development and testing of localization, mapping and control algorithms.
For indoor experiments, motion capture such as the Vicon~\cite{Delmerico2018, Merriaux2017} or Optitrack~\cite{Furtado2019} systems have become the \emph{de facto} standard.
They are based on tracking reflective or active markers by infrared cameras and offer an accuracy of sub-millimetre precision and high sampling rates.
However, the area that they can cover is limited by the number of cameras available.
Moreover, they struggle in direct sunlight.
Another high-fidelity source of reference pose is laser trackers based on laser interferometry, such as the one used by \citet{Sang2017}.
In exchange for the limitation of tens of meters in range, they offer position measurements accurate down to microns.
The possibilities of pose reference are more limited outdoors.
The approaches most prominent in the literature use \ac{GNSS}~\cite{Morales2007}, total stations~\cite{McGarey2018}, and qualitative comparison of trajectories to maps~\cite{Zhang2016}.
Besides, these outdoor referencing systems do not provide direct orientation measurements.

\begin{figure}[htbp]
	\centering
	\includegraphics[width=0.85\columnwidth]{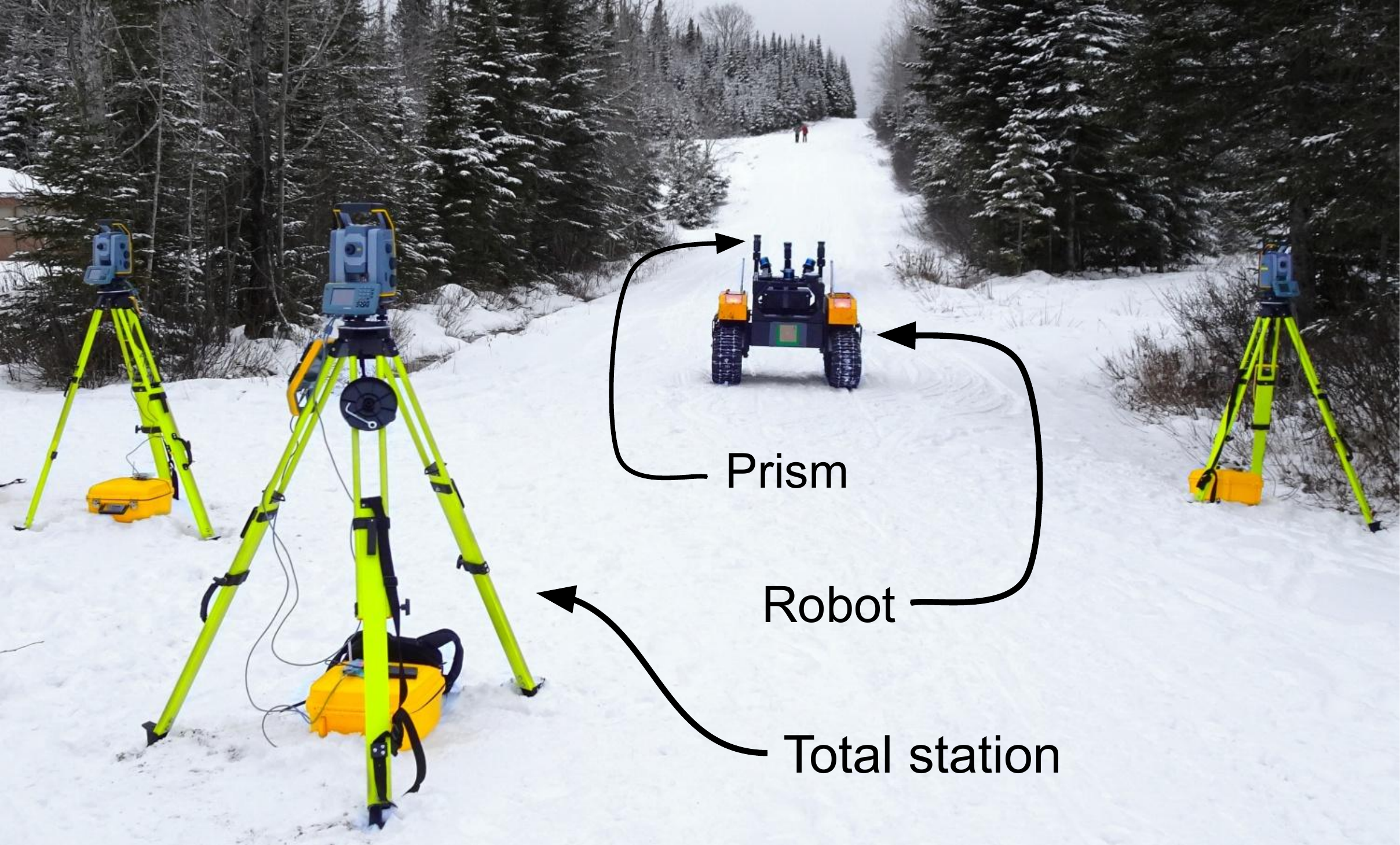}
	\caption{Example of an application where RTK positioning faces limitations.
	A robot navigating in a subarctic forest with three prismatic reflectors is shown. It is being tracked by three total stations.}
	\label{fig:intro}
\end{figure}

When it comes to gathering datasets in forests, such as the one depicted in \autoref{fig:intro}, accuracy of \ac{RTK} positioning suffers from signal attenuation by vegetation \cite{Kubelka2020}.
Consequently, the need for accurate and precise pose reference immune to foliage interference and signal reflection has motivated the development of a system based on multiple total stations appropriate for such task.
Here, we make the distinction between a theodolite, which is a precision optical measurement instrument used for measuring horizontal and vertical angles, and a total station, which is a theodolite with distance-measuring equipment.

In this article, we propose a protocol for a novel referencing system relying on multiple total stations to allow the measurement of the six \acp{DOF} of a vehicle.
More precisely, our contributions are :
\begin{enumerate}
    \item the calibration procedure of multiple total stations for a vehicle continuous tracking;
    \item the time synchronization and long-range data transmission protocol for online pose referencing;
    \item a precision analysis of our tracking procedure; and
    \item a comparison against a \ac{RTK} positioning system.
\end{enumerate}

%% file: tex/RW.tex
In the robotics literature, we find a large variety of position-referencing systems based on total stations.
They generally differ in the types of environment, in the number of referenced \ac{DOF}, or in the target robotic systems they are used for.

Regarding the robotic platforms, there are almost no limitations; the only one being how many prisms the robot can carry.
Researchers used total stations to track skid steered robots~ \cite{Krusi2015, MacTavish2018}, a tethered wheeled robot~\cite{McGarey2017}, planetary rover~\cite{Wettergreen2010, Helmick2004}, unmanned surface vessel~\cite{Hitz2015}, and walking robots~\cite{Sang2017, Bjelonic2018}.
In our application, we use a large skid-steered robot suitable for winter in subarctic forest.
However, the presented referencing system is generic and applicable to any robot that can carry three prisms distanced by more than \SI{80}{\cm}.

Tracking a robot in a forest using a total station was previously done by \citet{McGarey2018}.
They referenced the position of a wheeled tethered robot by a total station on steep rocky slopes in a forested area to validate the accuracy of localization and mapping algorithms.
However, total stations are also resilient to harsh environment, as they were used in deserts to track planetary rover~\cite{Helmick2004}.
Because of their long-range measurements, application on water bodies are possible.
For example, \citet{Hitz2015} used a total station installed on the shoreline to track the position of a robotic catamaran.
Total stations were also demonstrated in urban areas~\cite{Kubelka2015} as well as indoors~\cite{Lemus2014}.
Given their versatility in different environments, a referencing system relying on total stations is thus relevant for multiple applications.
The main limitation is the requirement of the line-of-sight between the prisms and the total stations. 

An important design choice is the number of \ac{DOF} the referencing system should track.
A total station naturally provides the position of a tracked target, thus three \acp{DOF}.
For example, in \cite{Kabzan2020} they mounted a prismatic reflector, or simply a prism, onto a race car to verify the output of its mapping and localization system.
Limited by their maximum payload capacity, system tracking \acp{UAV} are limited to one prism~\cite{Burri2016, Schmuck2019, Punzo2019}.
An interesting solution was presented by \citet{White2005}, which involves an actively-controlled pan-tilt reflector on a climbing inspection robot.
By always pointing the reflector towards the total station, two \acp{DOF} were added to the typical position measurement.
Finally, with at least three prisms on a robot, the full pose of a robot can be recovered using six \acp{DOF}.
In the work of \citet{Pomerleau2012}, a lidar dataset was collected by a sensor rig with ground truth information about the pose.
A single total station was used to manually measure three reflectors  while the rig remained static.
The same principle was used by \citet{Helmick2004} for a planetary rover.
We extend these works by proposing a system using multiple total stations to continuously track a vehicle, while it is moving, to recover its full pose.

%% file: tex/theory.tex
\subsection{Referencing system overview}

Total stations are better known for topographic surveys, but modern total stations offer the ability to track a target over one kilometre with a range precision within millimetres.
In order to acquire ground truth measurements, we propose using three total stations, each of them being labelled from one to three.
Each total station produces measurements in its own coordinate frame $\mathcal{F}_1, \mathcal{F}_2$ and $\mathcal{F}_3$.
We also propose to attach three prisms to the robotic platform, one per total station.
These prisms can be automatically detected by total stations in a \emph{tracking mode}.
In this mode, each total station continuously tracks a single prism and reports a measurements at a fixed rate.
Also, a unique electronic signature of each prism allows them to be distinguished while they are close to each other.

After installing the total stations on the experiment site, a calibration procedure is required to find the rigid transformation $\prescript{1}{2}{\mathbf{T}}$ from $\mathcal{F}_2$ to $\mathcal{F}_1$ and the rigid transformation $\prescript{1}{3}{\mathbf{T}}$ from $\mathcal{F}_3$ to $\mathcal{F}_1$.
This step is necessary to allow the expression of all total station measurements in a common coordinate frame, which is the one of total station 1.
Small reflective targets are positioned around the site in such a manner that they are visible to each of the total stations. 
The measurements are done using the \emph{single measurement} mode, which requires manually pointing the total stations to the targets.

Our setup requires that the data coming from all total stations be synchronized according to one central clock.
The clock chosen is the one on the embedded computer located on the robotic platform.
This conveniently allows time synchronization with the robot's own sensor measurements, while providing it with real-time pose updates. 
However, because of the long range tracking ability of the total stations, the distance between them and the robot can be high.
To solve this issue, long-range radio communication systems are necessary.

We aim for real-time data gathering and broadcasting instead of having files stored on the disk and post-processed later on.
Real-time tracking can allow the use of geo-fences for safety or the replacement of the on-board localization algorithm when testing other algorithms.
Each total station is connected to its own embedded computer with a wired USB connection. 
These embedded computers run a process which initiates the total station measurements and receives the raw data.
The raw data of the total station measurements are the vertical angle, horizontal angle, range to the prism, measurement time and the total station status.
These data are then marshalled by each of these embedded computers, dubbed \emph{clients}, via a long-range radio module over to the \emph{master} embedded computer, located in the mobile robotic platform.
The purpose of this master is to synchronize the data coming from the clients with a single clock and to control data flow between all long-range radios via a remote data collection protocol.
For broader acceptance and portability, we restrict ourselves to single-channel communication.
Because of this, the master must poll one client at a time to collect the data.
Our communication protocol is composed of two parts: a synchronization protocol between the master and the clients, and the sending of the data.

\begin{figure}[htbp]
	\centering
	\includegraphics[width=0.95\columnwidth]{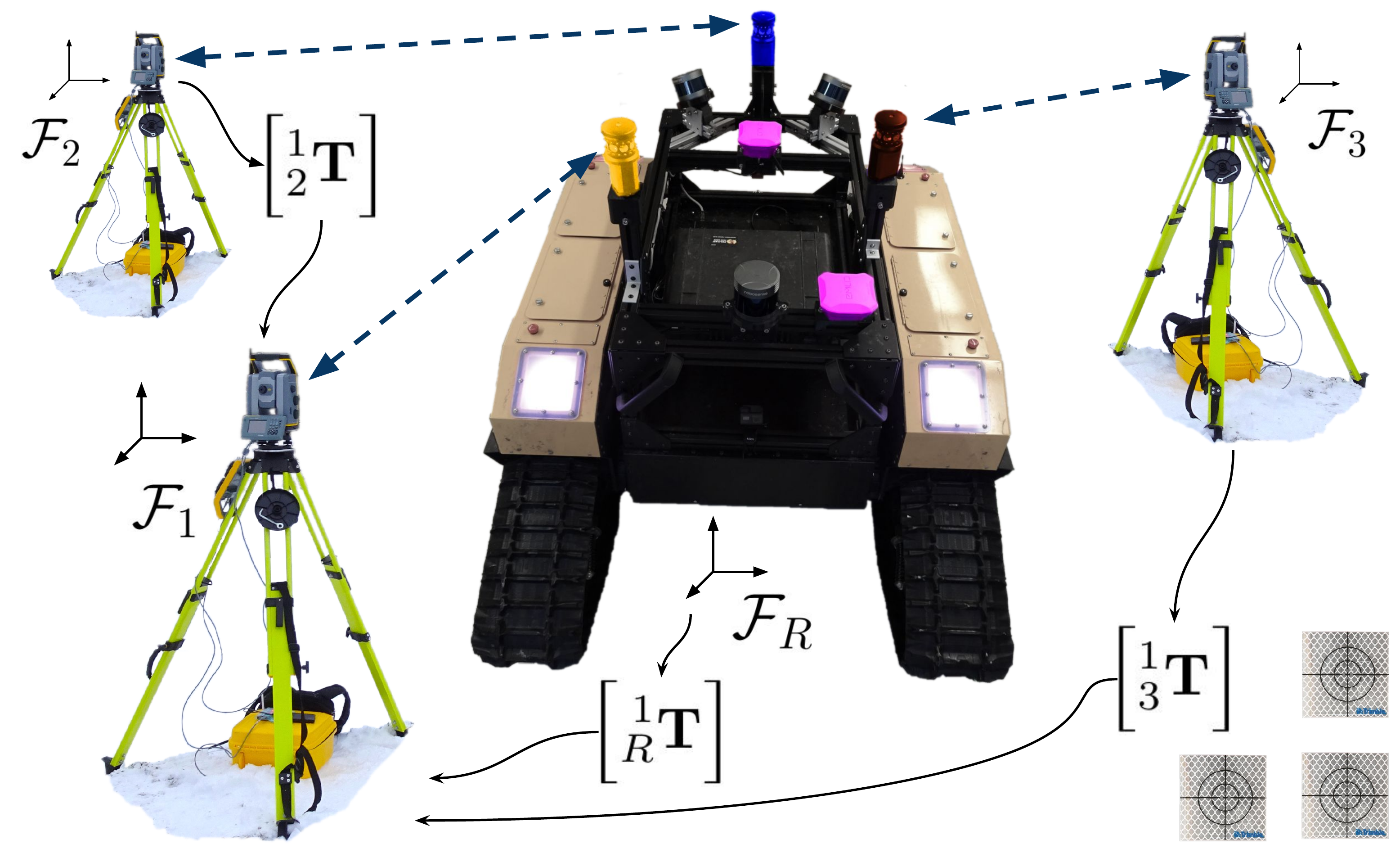}
	\caption{The experimental setup composed of three total stations, a Clearpath Warthog with prisms (golden, blue, sienna) and two GPS receivers (pink). The total station coordinate frames are related by the rigid transforms $\prescript{1}{2}{\mathbf{T}}$ and $\prescript{1}{3}{\mathbf{T}}$. We find these transforms by placing reflective targets (bottom right) around the experimental area and then measuring their position by all three total stations. The pose of of the robot is expressed by the transform $\prescript{1}{R}{\mathbf{T}}$ from the robot frame $\mathcal{F}_R$ to $\mathcal{F}_1$.}
	\label{fig:theodolite_setup}
\end{figure}

\subsection{Time synchronization}

\begin{figure}[ht]
	\centering
	\includegraphics[width=0.80\columnwidth]{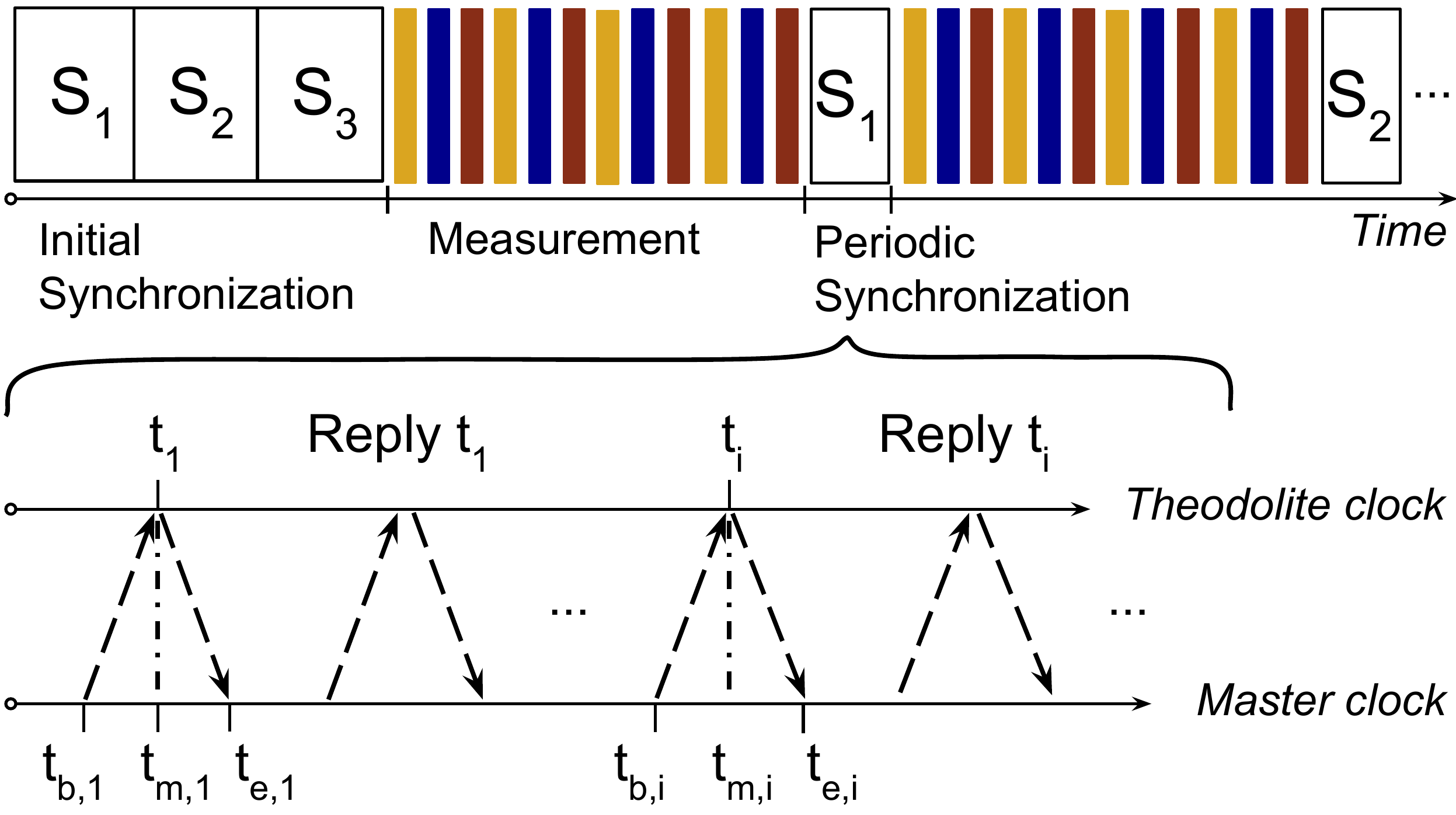}
	\caption{Communication protocol between the clients and the master radio module, with the detail of the time synchronization protocol.}
	\label{fig:timeline_lora}
\end{figure}

The system clocks on the clients are not synchronized with the system clock on the master. 
The synchronization protocol is used to find the skew between the clocks of the clients and the master.
To begin a synchronization, the master sends a message to ask a client to switch into this mode.
One client at a time can be synchronized.
The protocol is composed of cycles using four messages which is divided into two steps.
The number of cycles are set according to the configuration.
The general timeline of the communication protocol is shown in the top of \autoref{fig:timeline_lora}, and details about the synchronization procedure are in the bottom of the same figure.

First, two concurrent time stamps $t_{m,i}$ and $t_i$ are captured in the master and in the client.
The subscript $i$ is the cycle index, $m$ stands for \emph{middle}.
The two first messages are used for this purpose.
At time $t_{b,i}$ ($b$ as \emph{beginning}), the master sends a very short message (two bytes) to the client and saves the value of $t_{b,i}$.
Upon receipt of the message, the client stores the current time $t_i$ and also sends a very short reply to confirm the reception of the request. 
The master receives this confirmation at time $t_{e,i}$ ($e$ as \emph{end}), and records the value of $t_{e,i}$.
The timestamp $t_{m,i}$ can be estimated for every cycle as the average of $t_{b,i}$ and $t_{e,i}$. 

In the second step, we compute the skew between the master clock and the client clock.
By the third message of the cycle, the master requests the value $t_i$ from the client.
The last message of the cycle is the reply of $t_i$ value back to the master.
Using the value $t_{m,i}$, the skew of each cycle is found by subtracting the times $t_i$ and $t_{m,i}$.
Then, an average of the skew $\overline{d}$ is calculated over all $i$ cycles.
This average value is the correction which is applied on the timestamp of the data incoming from the corresponding client.

An initial synchronization is conducted at the beginning of the operation.
All of the clients are synchronized one at a time.
Then, the estimated skews are directly applied to correct the timestamps of the data.
Since the clients have a clock with significant drift, we need to synchronize them periodically during the operation.
To avoid losing too much measurement data, we only re-synchronize one client at a time during an experiment.
These re-synchronizations are done following the same protocol as during the initial synchronization, yet with less cycles.
Because less messages are sent to compute the time difference between the master and the client, the corrections calculated are not directly applied to the data timestamp.
A low-pass \ac{IIR} filter, which takes the former corrections $\delta_{j-1}$ calculated into account, is applied using
\begin{equation}
\label{eq:time_correction}
	\delta_j = w\overline{d} + (1-w)\delta_{j-1}.
\end{equation}
The final correction $\delta_j$ is applied after the $j^{th}$ synchronization. 
The filter weight $w$ is typically set $w \ll 0.5$ to slowly update the time offset.
This filter is used only after the initial synchronization.

After the synchronization protocol, the master switches the client to the normal measurement mode.
Following this protocol, the total stations can transmit the data, one total station at a time due to the single-channel available.
The master sends a measurement-request message to the total station in turn.
If a new measurement is available, it is sent back by the client. 
Otherwise, the master passes to the next total station and this communication cycle repeats itself.

\subsection{Calibration procedure}
\label{sec:calibration_frame}

To find the rigid transformations $\prescript{1}{2}{\mathbf{T}}$ and $\prescript{1}{3}{\mathbf{T}}$, a point-to-point alignment minimization is carried out by using the \ac{SVD} algorithm \cite{Arun1987} between the manually-measured reflective marker positions in the common frame $\mathcal{F}_1$ and the other two total station frames, $\mathcal{F}_2$ and $\mathcal{F}_3$.
Cartesian coordinates of each total station measurement can be projected into $\mathcal{F}_1$ using $\prescript{1}{2}{\mathbf{T}}$ and $\prescript{1}{3}{\mathbf{T}}$. 

It is also necessary to define the positions of the three prisms on the robot, with respect to the robot coordinate frame.
This measurement is performed accurately by one total station only once when mounting the prisms on the robot.
With the help of a \ac{CAD} software and the model of the robot, the prisms can be localized with respect to any chosen coordinate frame.

\subsection{Six-DOF pose estimation from prism reflector position measurements}
\label{sec:6dof_pose}

After calibration and synchronization of the system, we proceed to the actual measurements.
First, the validity of the data is checked.
If a total station reports a status indicating an error during the measurement (e.g., prism not detected, prism too close, total station not levelled), the measurement is not taken into account.
Once checked, the raw measurements are transformed into Cartesian coordinates in the corresponding total station frame and then into the common $\mathcal{F}_1$ frame.

\begin{figure}[ht]
	\centering
	\includegraphics[width=0.9\columnwidth]{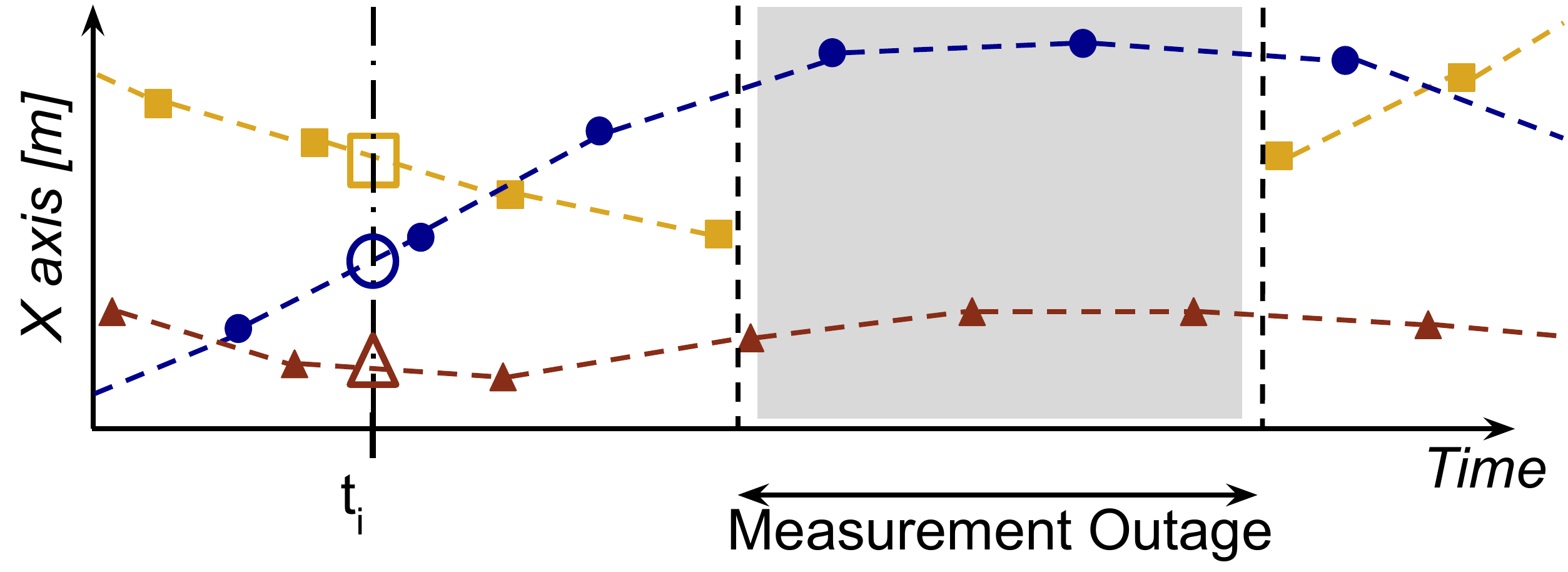}
	\caption{Example of measurement data interpolation for a single axis. Each color represents a single tracked prism. Solid shapes are actual measurements, while hollow shapes are interpolated estimations at time $t_i$.}
	\label{fig:interpolation}
\end{figure}

Once the prism position measurements all lie in the same frame, interpolation is performed as shown in \autoref{fig:interpolation}.
The figure shows the process for a single axis $X$, but the same is done for $Y$ and $Z$.
For a given time $t_i$, we interpolate the position of each of the three prisms between the actual measurements.
In the case of measurement outage of any total station for a time longer than a given threshold, the corresponding interval is marked as invalid for obtaining the ground truth reference.
An example of this situation is depicted in the grey interval of \autoref{fig:interpolation}. 
The chosen time step for interpolation is \SI{50}{\ms} to obtain a \SI{20}{\Hz} sampling rate in the reference.
Results are triplets of coordinates corresponding to the positions of the three prisms at time $t_i$.

For each triplet, a point-to-point alignment minimization is performed, using the same method as the one used during the $\mathcal{F}_1$-$\mathcal{F}_2$-$\mathcal{F}_3$ calibration procedure.
The problem can be summarized as minimizing the point-to-point cost function $\mathrm{J}_\text{p-p}(\cdot)$ as
\vskip -3mm
\begin{align}
\label{eq:icp_p_to_point_min}
\prescript{1}{R}{\mathbf{T}} = \arg \min_{\bm{T}} \: {J}_\text{p-p} \left( \mathcal{Q}, \mathcal{P} \right)  
\text{, \hspace{0.05cm} with \hspace{0.05cm}}
{J}_\text{p-p} = \sum_{k=1} (\bm{e}_{k}^{T} \bm{e}_k)\text{,}
\end{align}
\vskip -2mm
where $\bm{e}_{k} = \bm{q}_k - \bm{T}\bm{p}_k $ is the error vector between the $k$\textsuperscript{th} point $\bm{q}$ of $\mathcal{Q}$ representing the measured prism positions taken as reference, and the $k$\textsuperscript{th} point $\bm{p}$ of $\mathcal{P}$ representing the positions of the prisms as mounted on the robot and calibrated in the lab, and $\bm{T} \in \mathrm{SE}(3)$ is a rigid transformation matrix used during the minimization.
The result given by \ref{eq:icp_p_to_point_min} is a rigid transformation matrix $\prescript{1}{R}{\mathbf{T}} \in \mathrm{SE}(3)$ between the frame $\mathcal{F}_1$ and the frame of the robot $\mathcal{F}_R$, which gives the 6-\ac{DOF} pose of the robot body frame (a translation vector and a rotation quaternion) in the common frame $\mathcal{F}_1$.

%% file: tex/experimentations.tex
\begin{figure}[htbp]
	\centering
	\includegraphics[width=0.555\columnwidth]{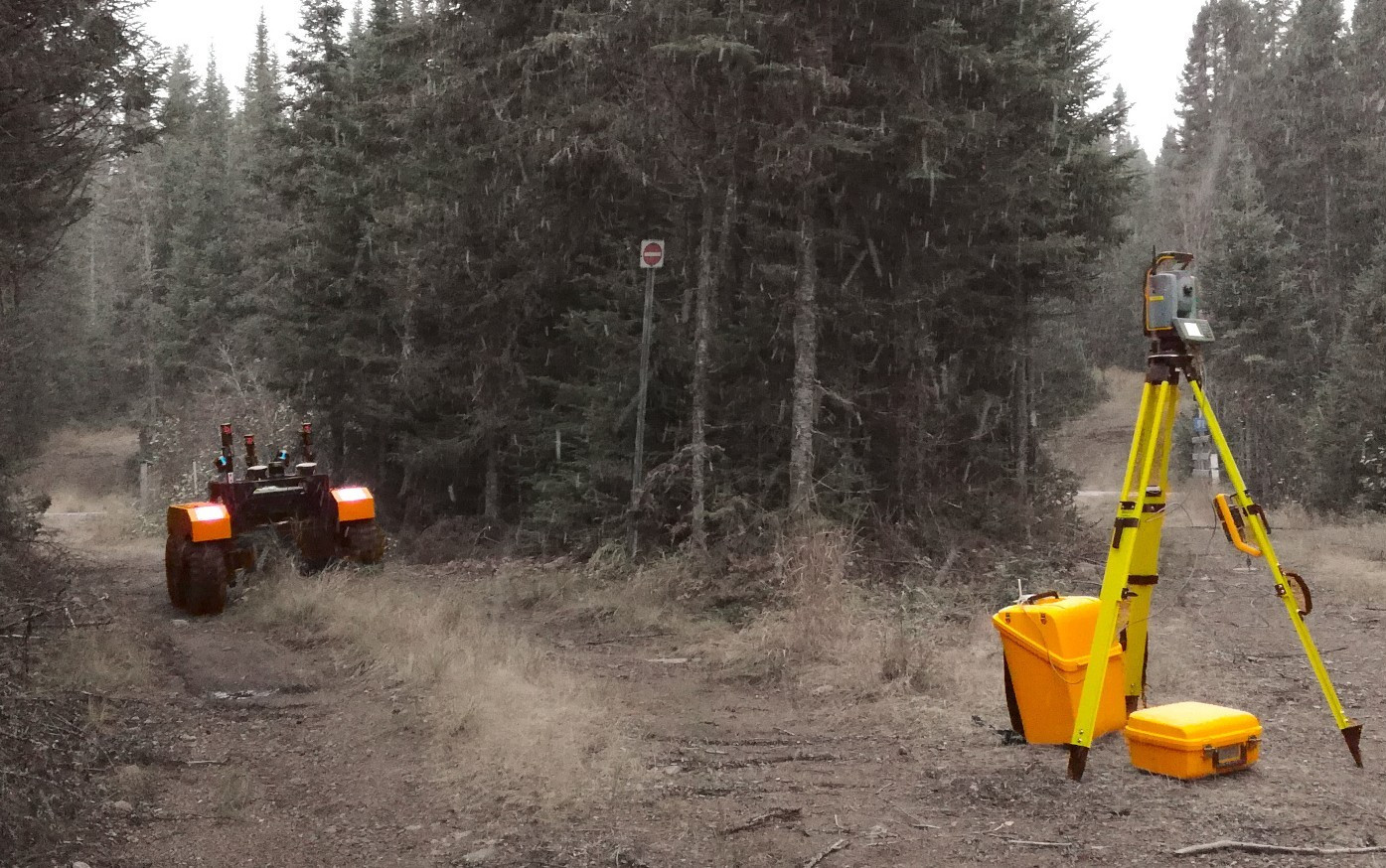}
	\includegraphics[width=0.425\columnwidth]{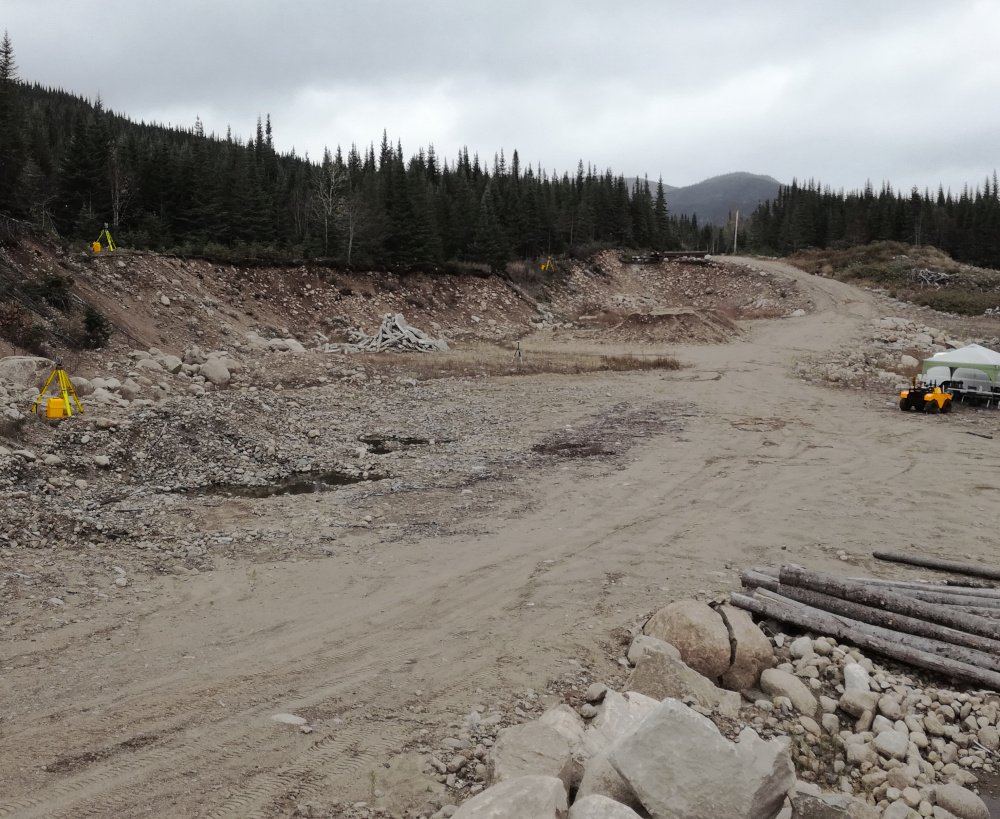}
	\caption{Experimental environments.
	\emph{Left}: the ski trail. \emph{Right}: the stone quarry areas.}
	\label{fig:ski-trail}
\end{figure}

To evaluate the performance of our position referencing system, we conducted several experiments in different locations around the Montmorency Forest.
Later is a research facility belonging of Laval University, located north of Quebec City.
The specific locations were \emph{i)} one experiment in a former quarry , \emph{ii)} three experiments in a forest cross-country ski trail, and \emph{iii)} one experiment in a forest snowmobile trail.
\autoref{fig:intro} and \autoref{fig:ski-trail} show these environments.
These locations were chosen to test our setup in different operating conditions: medium path of around \SI{200}{\m} in an open space for the quarry, short path of around \SI{150}{\m} with foliage and trees on the ski trail, and longer path of around \SI{600}{\m} in the forest on the snowmobile trail.
Hence, we expect some \ac{GNSS} signal attenuation and some long-range radio signal attenuation in the forest area.
These will allow us to evaluate our total stations setup compared to the accuracy of a \ac{GNSS} setup, but also to evaluate the tracking mode of the total station over different experiments.
The evaluations will compare the results obtained during each of these experiments.

\subsection{Equipment used}

The particular equipment used for our experiments is displayed in \autoref{fig:theodolite_setup}.
In this figure, we can see a \emph{ClearPath Warthog} mobile robot used as the test robotic platform, along with three \emph{Trimble MultiTrack} prisms on the platform.
These active prisms emit an infrared signature producing a unique identifier to avoid confusion during tracking.
They were individually tracked by three total stations \emph{Trimble S7}.
In nominal conditions, the distance measurement has an accuracy of \SI{2}{\mm} and the range of the tracking mode is \SI{800}{\m}.
The angular accuracy is \SI{1}{''} for the vertical and horizontal angles.
This means that at \SI{800}{\m}, the angular accuracy will affect the precision by an error of almost \SI{4}{\mm}.
The maximum achievable measurement rate of a \emph{Trimble S7} in the prism-tracking mode is \SI{2.5}{\Hz}.

The radio communication between all of the total stations and the master computer is provided by four \emph{Raspberry Pi 4} computers with \emph{LoRa Shield} radio modules from \emph{Dragino}.
The parameters used for the communication are listed in \autoref{tab:params}.
The chosen modulation allows the Raspberry Pi to reliably send data over a distance of up to \SI{800}{\m} at 366 {Bps} in open space.
With our communication protocol, we can poll data from all three total stations at a frequency of approximately \SI{1.4}{\Hz}.

\begin{table}[htbp]
	\centering
	\caption{Table of the radio protocol communication parameters used.} 
	\label{tab:params}
	\begin{tabu}{X[2,c]X[1,c]}
		\toprule
		\emph{Parameters}& \emph{Values} \\
		\midrule
		LoRa frequency & \SI{905}{\MHz}\\
		Spreading factor & 9\\
		Error coding rate & 4:6\\
		Maximum byte rate & \SI{366}{Bp\second}\\
		Measurement message length & \SI{34}{B} \\
		Low-pass filter parameter $w$ & 0.1\\
		Initial synchronization cycles & 50\\
		Re-synchronization cycles & 5\\
		Measurement outage threshold & \SI{1}{\s}\\
        \bottomrule
	\end{tabu}
\end{table}

Three \emph{Reach RS+} \ac{RTK} receivers were used for comparison purposes, two on the robot and one as a reference on a tripod.
The distance between each prism, as well as the distance between the two \ac{GNSS} receivers on the robot were accurately measured by a single total station in the lab.
The obtained values are shown in the \autoref{tab:distance}.
To assess the accuracy of our localization approach, we compare the reference values from \autoref{tab:distance} and the ones measured during the experiments for noise evaluation, similarly to the work of \citet{Pomerleau2012}.

\begin{table}[htbp]
    \centering
    \caption{Distance between the prisms and between the \ac{GNSS} receivers, used as reference measurements.} 
    \label{tab:distance}
    \begin{tabu}{X[1,c]X[1,c]X[1,c]X[1,c]X[1,c]}
    \toprule
     & \emph{Prisms 1-2} & \emph{Prisms 1-3} & \emph{Prisms 2-3} & \emph{GPS 1-2}\\
    \midrule
    \emph{Distance} & \SI{987}{\mm} $\pm$ \SI{2}{\mm} & \SI{681}{\mm} $\pm$ \SI{2}{\mm} & \SI{815}{\mm} $\pm$ \SI{2}{\mm} & \SI{810}{\mm} $\pm$ \SI{2}{\mm}\\    
    \bottomrule
    \end{tabu}
\end{table}

\subsection{Evaluation in static conditions}

\begin{figure}[htbp]
	\centering
	\includegraphics[width=\columnwidth]{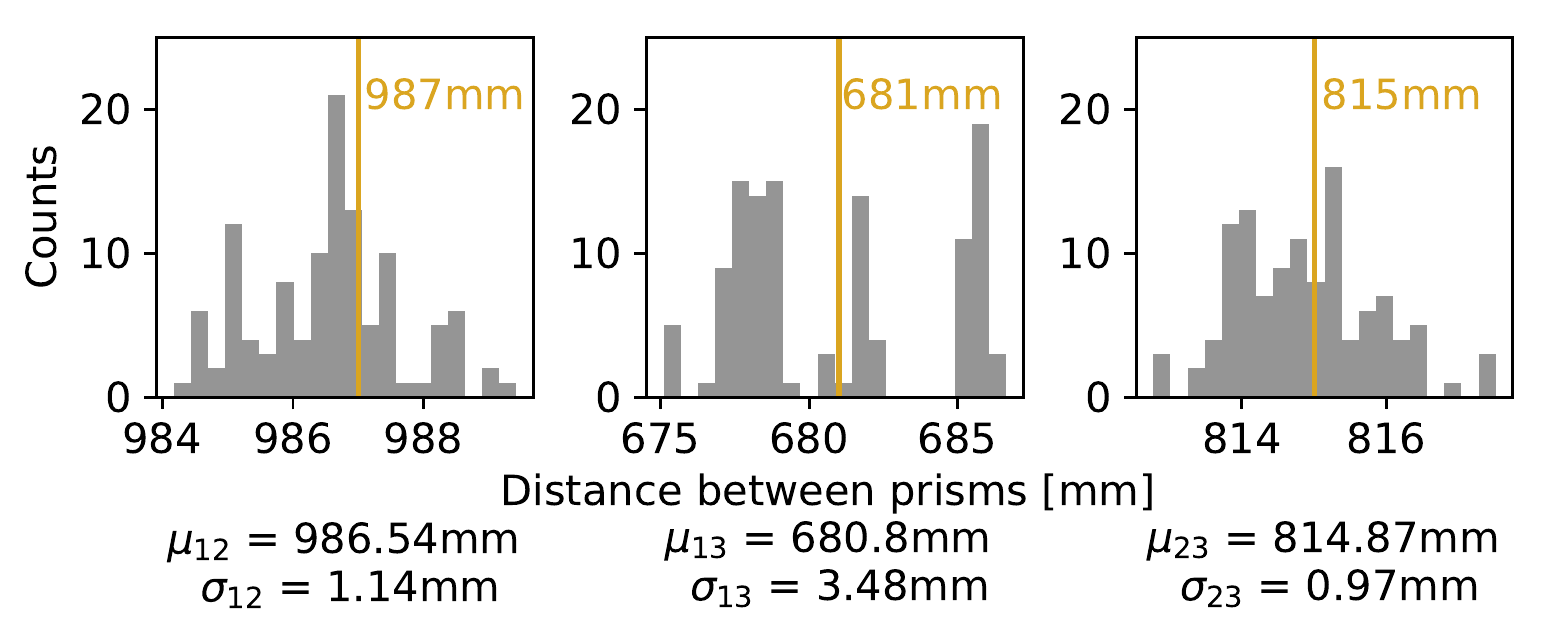}
	\caption{Histogram of the distance between prisms during stationary poses. 
	The golden lines represent the reference distance between the prisms given in \autoref{tab:distance}.}
	\label{fig:histogram_ski_trail}
\end{figure}

We first evaluate the precision of our system with the robot during stationary poses from data taken at the ski trail shown in \autoref{fig:ski-trail}-\emph{left}.
We identified the intervals when the robot was stationary based on the recorded odometry.
For these intervals, the distances between the measured prism positions were calculated.
We used the raw data of the total stations without the interpolation for this purpose.
This was done to evaluate the accuracy of the calibration procedure, which does not depend on the interpolation, and also to evaluate the precision of the total stations themselves.
The 345 resulting distance samples are aggregated in the histograms shown in \autoref{fig:histogram_ski_trail}.
We can compare the mean values $\mu_{12}$, $\mu_{13}$, and $\mu_{23}$ with the reference values from the \autoref{tab:distance}, shown as the golden lines in the the histograms.
The absolute errors are respectively \SI{0.46}{mm}, \SI{0.2}{mm}, and \SI{0.13}{mm}.
The precision is high with the maximum standard deviation of \SI{3.5}{\mm} in the case of $\sigma_{13}$.
This is in line with the expected precision of the total stations using the tracking mode with a prism, given the \SI{2}{\mm} RMSE specification from the manufacturer (\emph{Trimble}).
This also confirms that the calibration procedure used to compute the common frame $\mathcal{F}_1$ was accurate enough.
We can also observe a symmetrical result around the true value of the inter-prism distance 1 to 3, which results in a larger dispersion $\sigma_{13}$ of the data than the $\sigma_{12}$ and $\sigma_{23}$ dispersion of the data.
We believe this is due to the calibration precision between the total station frames, which caused a small error in rotation in the transform matrix $\prescript{1}{3}{\mathbf{T}}$.
This is due to the experiment itself, where the robot moved towards and then away from the total stations.

\subsection{Evaluation in dynamic conditions}

\begin{figure}[htbp]
	\centering
	\includegraphics[width=0.95\columnwidth]{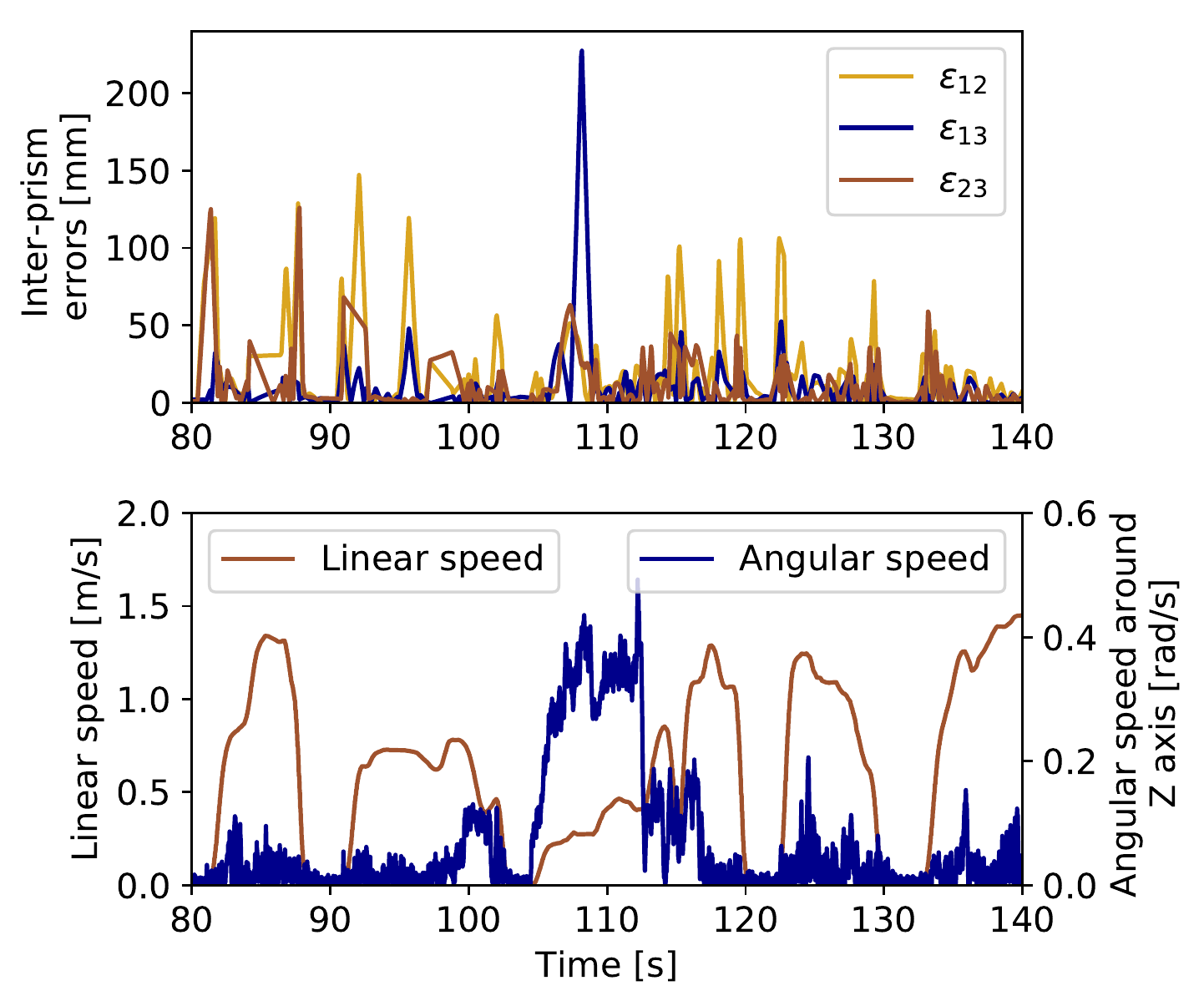}
	\caption{Results in dynamic conditions in the ski trail experiment.
	\emph{Top}: the inter-prism errors.
	\emph{Bottom}: the corresponding linear and angular speed of the robot.}
	\label{fig:distance_error_prims_time}
\end{figure}

Next, we evaluate the performance during the motion of the robot. 
\autoref{fig:distance_error_prims_time} shows the inter-prism error over time obtained from one ski trail experiment.
During this experiment, the robot was driven in a start-and-stop motion fashion.
The plot shows a representative portion of the run, between \SI{80}{\s} and \SI{140}{\s}, as the rest of the run was similar.
The linear speed and angular speed of the robot during this experiment are shown in the bottom part of the figure.
These graphs show that the reference is less precise when the robot starts to move, or when it stops.
At these moments, the error increases, leading to the several spikes in the distance error plot.
Also, there is an increase in the error when the robot turns at time \SI{109}{\s}. 
We attribute this error to the ability of the total stations to accurately track the prisms during abrupt changes of velocity.
Also, the sampling scheme dictated by our hardware has an influence on the precision in this dynamic case.
One needs to recall that the system has a relatively low sampling rate and the measurements of the three prisms are not taken synchronously.

\begin{figure}[!ht]
	\centering
	\subfloat[][The error according to the linear and angular speed.]{
	\includegraphics[width=0.45\textwidth]{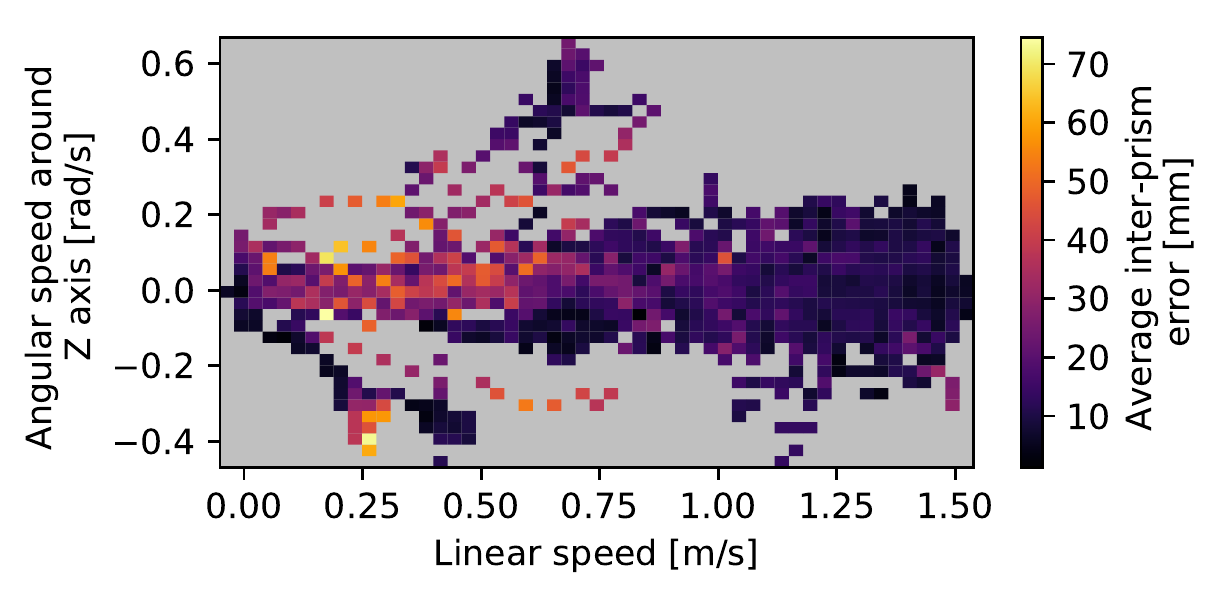}
	\label{fig:grid_linear_angular}}
	\hfill
	\subfloat[][The error according to the acceleration and angular speed.]{
	\includegraphics[width=0.45\textwidth]{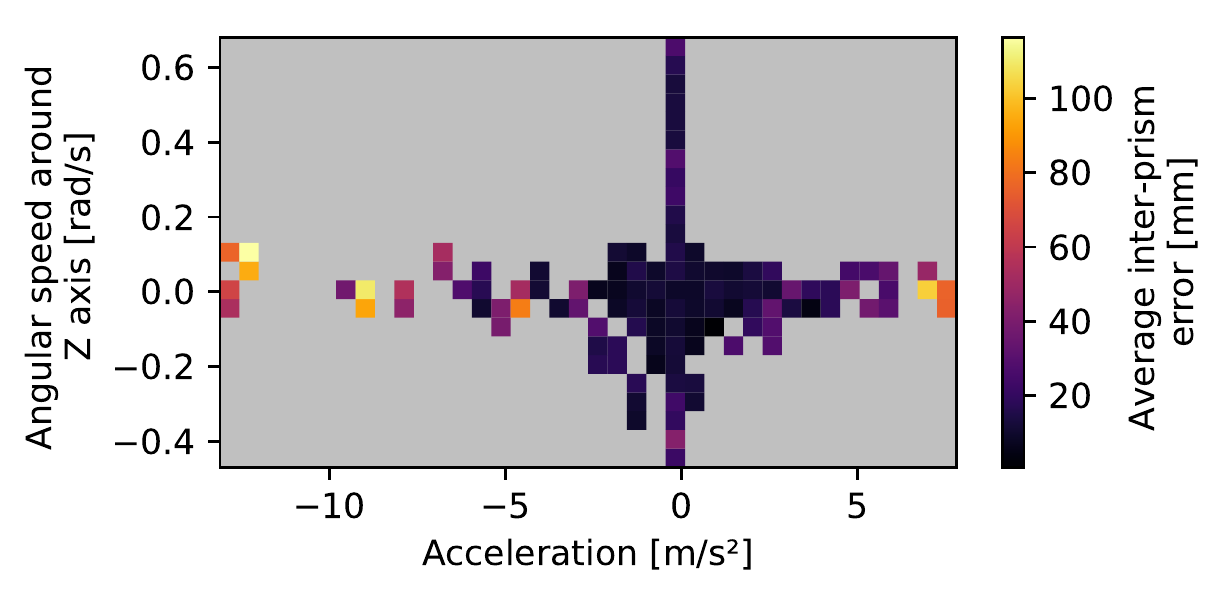}
	\label{fig:grid_accel_angular}}
	\caption{The distance error between prisms during all ski trail experiments.}
	\label{fig:grid_plot}
\end{figure}

To examine the influence of the linear speed, acceleration and angular speed, all data taken during the three ski-trail experiments were gathered together.
To each inter-prism distance error, the values of linear speed, acceleration and angular speed were attributed, as depicted in \autoref{fig:grid_plot}.
All of the samples belonging to a single bin were averaged.
The first subplot (a) represents the inter-prism distance error according to the linear speed and the angular speed of the robot.
We see a higher error at linear speeds between \SI{0.25}{\m/\s} and \SI{0.5}{\m/\s}. %
The error also increases with the angular speed.
We think that these results have a correlation with the way a human operator drives the robot.
The robot driven by the operator either goes straight forward at higher velocity or turns in lower velocities. 
This human bias is also why the angular velocities better represented in the plot at low linear speeds.
When moving straight forward, our sampling rate is sufficient for the interpolation to return correct values.
However, when the robot turns, we can miss some key samples and thus obtain less precise results.
The second subplot (b) shows the inter-prism distance error as a function of acceleration and angular speed.
The error spikes due to the start-stop motion shown in the \autoref{fig:distance_error_prims_time} are clearly highlighted at both extremes of the acceleration axis.

\subsection{From prism uncertainties to pose uncertainty}

\begin{figure}[htbp]
	\centering
	\includegraphics[width=\columnwidth]{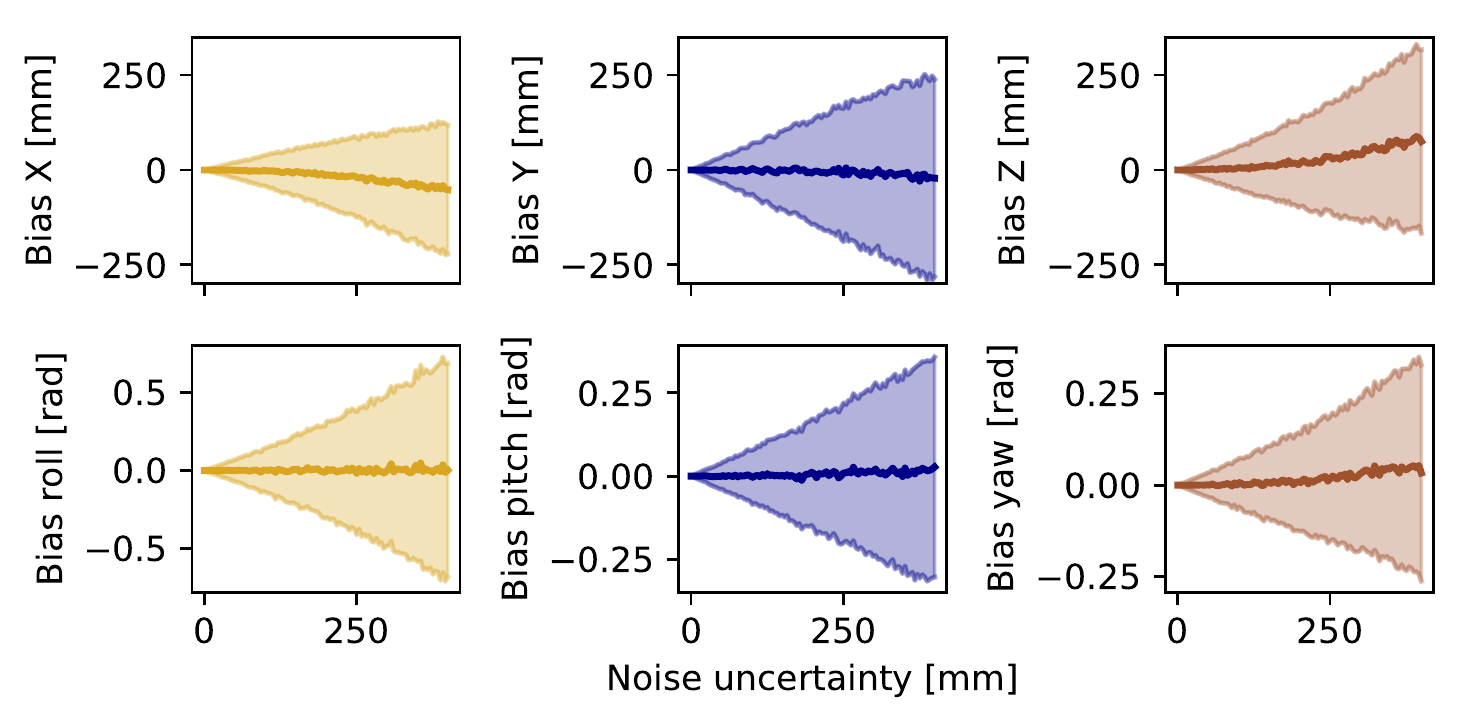}
	\caption{Perturbation analysis of the 6-DOF pose estimation from the prism measurements.
	The solid line and the shaded area represent respectively the mean value and the standard deviation.
	}
	\label{fig:noise_uncertainty}
\end{figure}

We analyze the impact of uncertainty in the prism measurements on the measured robot position and orientation.
As explained in \autoref{sec:6dof_pose}, after the interpolation, a point-to-point alignment minimization is performed to obtain the 6-\ac{DOF} pose of the robot frame.
To determine the level of uncertainty in the 6-\ac{DOF} pose that results from the noise in the total station measurements, we carried out the following perturbation analysis: a Gaussian white noise was randomly applied to the positions of the prisms with the standard deviation range of \SI{0}{\mm} to \SI{400}{\mm} with a step of \SI{4}{\mm}.
This approach was used to simplify the study of the noise propagation through the point-to-point minimization, which uses non-linear functions.
The range used is comparable to the uncertainty that is observed during our experiments. 
For each value in the range of standard deviation, 1000 simulations of point-to-point alignment minimization were performed.
The resulting sets of 1000 poses allowed us to study the distribution of position and orientation errors.
\autoref{fig:noise_uncertainty} shows these pose errors as a function of the input perturbation.
The mean and standard deviation are extracted and plotted for each axis in position and orientation.
In this analysis, we choose the Euler angles representation of the orientation in \autoref{fig:noise_uncertainty} for the simplicity of their interpretation.
The values are fairly low and thus far from the singularities in this representation.
Not surprisingly, the spread of the inter-prism error position is very close to a linear function depending on the noise uncertainty.
The same applies for the Euler angles, at least in the noise range we have chosen.
Interestingly, the mean value is not zero.
Its absolute value is slightly increasing as a non-linear function of noise range.
Nevertheless in our experiments, the mean value of the inter-prism measurement error is \SI{10}{\mm} and the vast majority of its distribution lies below \SI{50}{\mm}.
In this region, the observed bias in the resulting position and orientation can be neglected. 
Moreover, the mean value of the inter-prism measurement error of \SI{10}{\mm} translates to \SI{10}{\mm} in position and \SI{0.01}{rad} (\SI{0.6}{\deg}) in orientation, based on sampling from an equivalent input error distribution.

\subsection{Comparison with \ac{GNSS} solutions}
\label{sec:GPS_analysis}

To ground the performance of the total station referencing system to a commonly-known alternative, we compared it to the \ac{RTK} solution.
In our previous work \cite{Kubelka2020}, we have studied the impact of vegetation and obstacles both on the propagation of the satellite signal and the base station correction signal.
The satellite localization performed well in open areas, but a decrease in accuracy was observed when portions of the sky view were obstructed by vegetation.
To this effect, we have recorded the \ac{RTK} positioning data during two experiments: one in the quarry and another one in the forest.
The raw \ac{GNSS} data were recorded as log files in the receivers and the \ac{RTK} solution was obtained in post-processing by the means of the \emph{RTKLIB}\footnote{\url{http://www.rtklib.com/}} library.
We selected the \ac{RTK} solver settings that would be achievable in real-time operation (forward-pass filtering only). 

\begin{figure}[!ht]
	\subfloat[][Inter-GPS error during the quarry experiment without RTK solution.
	The results obtained are \{$\mu_{GPS} = \SI{915}{\mm}$, $\sigma_{GPS} = \SI{888}{\mm}$\}.]{
	\includegraphics[width=0.49\textwidth]{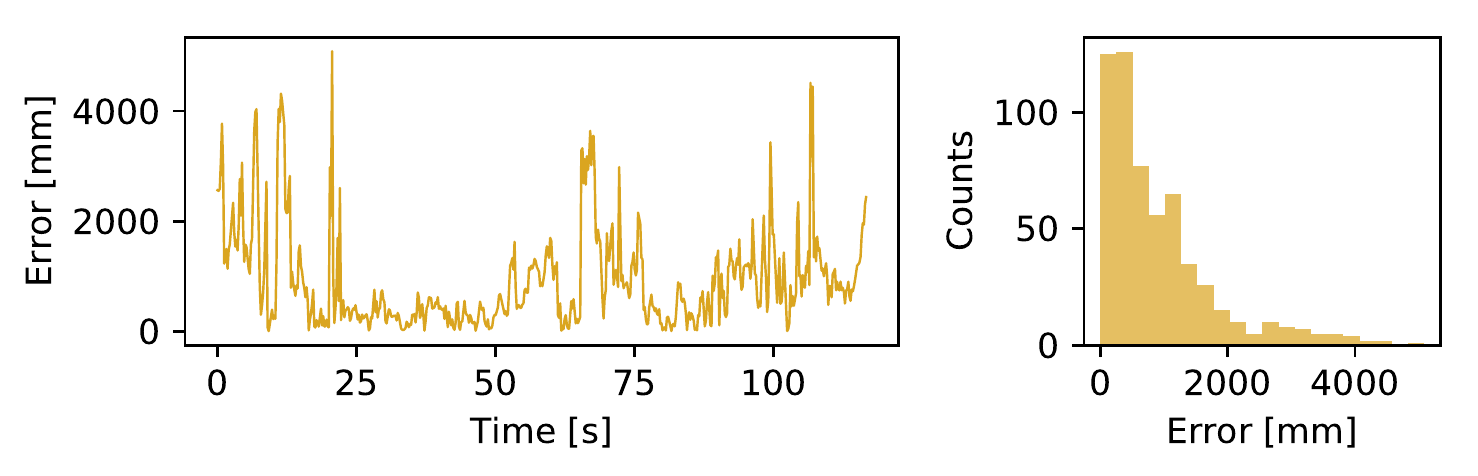}
	\label{fig:gps_basic_quarry}}
	\hfill
	\subfloat[][Inter-prism and inter-GPS error during the quarry experiment with the RTK solution.
	The results obtained are respectively \{$\mu_{Theo} = \SI{9.6}{\mm}$, $\sigma_{Theo} = \SI{8.6}{\mm}$\}  and \{$\mu_{RTK} = \SI{10.2}{\mm}$, $\sigma_{RTK} = \SI{5.0}{\mm}$\}.]{
	\includegraphics[width=0.49\textwidth]{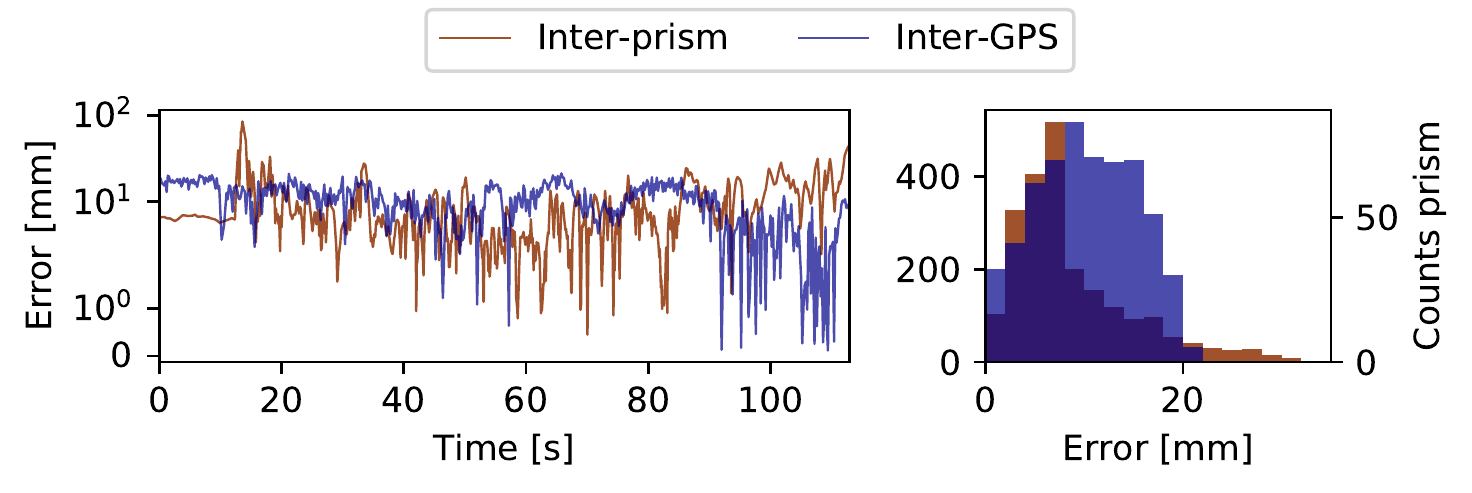}
	\label{fig:gps_rtk_quarry}}
	\hfill
	\subfloat[][Inter-prism and inter-GPS error during the forest experiment with RTK solution.
	The results obtained are respectively \{$\mu_{Theo} = \SI{7.0}{\mm}$, $\sigma_{Theo} = \SI{6.2}{\mm}$\} and \{$\mu_{RTK} = \SI{496}{\mm}$, $\sigma_{RTK} = \SI{622}{\mm}$\}.]{
	\includegraphics[width=0.49\textwidth]{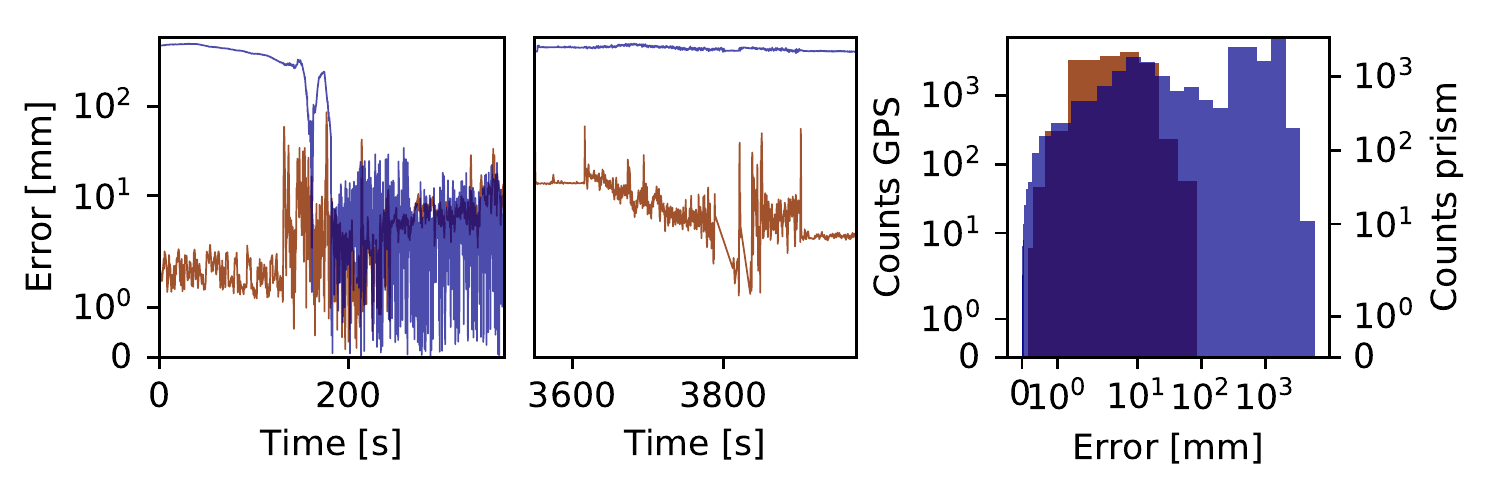}
	\label{fig:gps_rtk_forest}}
	\caption{The distance errors of our system compared to GPS during the quarry and the forest experiments.}
	\label{fig:gps_distance}
\end{figure}

Similarly to using the inter-prism distance to estimate the total station localization precision, we rely on the measured distance between the \ac{GNSS} receivers to estimate their precision. The distance between these \ac{GNSS} receivers is indicated in \autoref{tab:distance}.
\autoref{fig:gps_distance} shows the results from the quarry experiment and from the snowmobile trail experiment.
The purpose of these experiments is to compare the results coming from an open-area and from a more obstructed one.
The average number of satellites used by the receivers during the quarry experiment was $15.3\pm1.4$ (minimum 2 and maximum 20 satellites), and $13.1\pm2.8$ satellites in the forest area during the snowmobile trail experiment (minimum 3 and maximum 21 satellites).
For reference, we show the precision of the basic \ac{GNSS} solution without the \ac{RTK} corrections in \autoref{fig:gps_distance}~(a).
Even in the open area of a quarry, the precision is essentially unable to accurately locate the robot.
In a more realistic comparison, \autoref{fig:gps_distance}~(b) compares the precision of the corrected \ac{RTK} localization and our total station referencing system.
In this environment and with the robot moving, the precision is equivalent as shown on the histogram.
The total station system has a slightly higher error spread compared to the \ac{RTK} solution due to the error spike in the beginning of the experiment.
This error arises from an incorrect inter-prism distance 1-2, probably due to the acceleration of the robot when it begins to move, similarly as in \autoref{fig:distance_error_prims_time}.
Otherwise, the accuracy is similar for both.
In a more difficult scenario, \autoref{fig:gps_distance}~(c) shows portions of the snowmobile trail experiment, where the robot was in reach of all three total stations.
From the beginning until the mark of \SI{200}{\s} and in the remainder of the experiment (right plot), the \ac{GNSS} signals are significantly attenuated by the trees growing on both sides of the trail.
During this part, the \ac{GNSS} precision drops drastically, hovering around \SI{1.6}{\m}.
From the mark \SI{200}{\s}, the sky view opens and the \ac{GNSS} achieves its nominal precision again.
On the other hand, the average precision of the total station reference is \SI{7.0}{\mm} as long as the robot lies in the field of view of the total stations.

These two experiments show that the proposed total station setup has an equivalent accuracy and precision compared to the \ac{RTK} solution in an open area.
However, our solution is significantly better in an obstructed environment such as a forest, where the \ac{GNSS} signal is weaker because of the vegetation.
Our setup can thus be used as a source of precise ground truth in such an environment.

%% file: tex/conclusion.tex
In this paper, we proposed a referencing system based on total stations in order to provide 6-\ac{DOF} pose measurements in real time.
The referencing system was demonstrated successfully on a large tracked skid-steered robot in a subarctic forest. 
With the robot not moving, the precision achieved was better than \SI{1}{\mm} in position and \SI{0.06}{\deg} in orientation.
These values are in line with the limits of the total stations used for the experiments.
With the robot moving, the precision decreased to \SI{10}{\mm} in position and \SI{0.6}{\deg} in orientation.
In this dynamic case, the other factors are the ability of the total stations to quickly react to speed changes and the measurement rate, which may be hiding the dynamics.
Nevertheless, the experiments have demonstrated the robustness of our results against different types of environments.
Also, they are equivalent to the results obtained by the \ac{RTK} solution in open areas and by a large margin better in environments where \ac{GNSS} signals are weak.

In future work, the reflective calibration markers will be replaced by a dedicated, automatically tracked prism.
This will increase the inter-total station calibration accuracy and significantly decrease the setup time.
We will study in more detail the effect of acceleration of the robot, which we consider to be the main source of error in our experiments at the moment.
A similar problem is addressed by \citet{Ehrhart2017}, where they propose improvements based on data from an RGB camera.
This could be a starting point to address this limitation.
The total stations can achieve the maximum measurement rate of \SI{2.5}{Hz}.
We will further optimize the long-range communication protocol to fully use this (limited) potential.
Finally, we have to assess the influence of prism configuration on the robot on the uncertainty of the final pose, and possibly finding an optimal configuration.